# IMPORTANCE OF IMAGE ENHANCEMENT TECHNIQUES IN COLOR IMAGE SEGMENTATION: A COMPREHENSIVE AND COMPARATIVE STUDY

DIBYA JYOTI BORA[a]

[a] Department of Computer Science and Applications Barkatullah University, Bhopal, India

[a] Email: research4dibya@gmail.com

## ABSTRACT

Color image segmentation is a very emerging research topic in the area of color image analysis and pattern recognition. Many state-of-the-art algorithms have been developed for this purpose. But, often the segmentation results of these algorithms seem to be suffering from miss-classifications and over-segmentation. The reasons behind these are the degradation of image quality during the acquisition, transmission and color space conversion. So, here arises the need of an efficient image enhancement technique which can remove the redundant pixels or noises from the color image before proceeding for final segmentation. In this paper, an effort has been made to study and analyze different image enhancement techniques and thereby finding out the better one for color image segmentation. Also, this comparative study is done on two well-known color spaces HSV and LAB separately to find out which color space supports segmentation task more efficiently with respect to those enhancement techniques.

# I. INTRODUCTION

Color images carry a vast amount of information with them. But this information is somewhat hidden, so human eyes tend to fail in analyzing them. Most importantly, small changes in characteristics of information such as intensity, color, texture etc are really difficult to get realized. So, we need an efficient color image segmentation technique to analyze them. But the result of any color image segmentation technique totally depends on the quality of the image concerned. Especially, in the case of the satellite image, image quality is degraded because of noises that generally involved during capturing, transmission and acquisition process of the image. So, segmenting such noisy images does not produce an effective analysis result. Hence, we need some preprocessing techniques to remove artifacts, outliers or we can say noises from the images before going for further analysis stage. Image enhancement is such a preprocessing technique where our goal is to suppress the noise while preserving the integrity of edges and the other detailed information (Gonzalez et al.,2014; Malik et al.,2014 ). Actually, noises can be removed completely only when the real causes of their formation are studied and investigated. But in real fact, we cannot completely investigate them. So, the only thing we can do is to introduce some mathematical equation based techniques to partially remove the noises as much as possible (Koschan et al, 2008).

Color image enhancement techniques involve more efforts than gray image enhancement techniques due to the following two reasons (Koschan et al, 2008):

(1) In the case of color images, we need to consider vectors instead of scalars.

(2) Also, for color images, the complexity of image perception is again a considerable fact.

Vector based techniques are computationally very hard to implement. So, monochromatic based techniques are always given preference where separate channels of a color image are undergone enhancement. Color space always matters a lot when it comes to color image processing. RGB is the common one when we talk about color image. This color space has three components Red, Green, and Blue. So, monochromatic based segmentation involves analyzing these three channels and we cannot guarantee a very good result for this color space. For this reason, next option to go for those color spaces where we have a separate channel for brightness measurement. HSV and LAB are two most popular color spaces satisfying this criterion. In this paper, we have adopted these two color spaces for our experimental study. A detail illustration on LAB color space can be found in (Bora et al, September 2014) and (Bora et al, December 2014) explains about HSV color space.

The later portion of the paper is organized as follows: In section II and its subsections, we have investigated different image enhancement techniques for color images. In section III, a brief discussion on color image segmentation is presented. In section IV, a study on color space is given where two frequently adopted color spaces HSV and LAB are explained properly with real-time examples. Section V is the experiments and results in the discussion section. Finally, the conclusion is drawn in Section VI.

## II. CONTRAST ENHANCEMENT

Contrast enhancement is a process by which the pixel intensity of the image is changed to utilize the maximum possible bins (Gupta et al, 2014). Generally, the "contrast" term refers to the separation of dark and bright areas present in an image. The advantage of contrast enhancement is that it removes the ambiguity that may otherwise arise between different regions in an image. Contrast enhancement can be categorized into two categories: (1) Local contrast enhancement; and (2) Global contrast enhancement. Following table illustrates them more precisely:

|   | **Local Contrast Enhancement** | **Global Contrast Enhancement** |
|---|---|---|
| 1. | In this type of contrast enhancement techniques, a small window is a slide through every pixel of the input image sequentially and only those blocks of pixels are enhanced which fall under in this window. This means local information is used intelligently in this case. | In this type of contrast enhancement techniques, the global histogram information is considered for enhancement. As the whole image is considered at once, so, local information is ignored in this case. |
| 2. | Local brightness future is considerably improved in this case. So contrast ratio can be improved in every region of the image. | As global brightness future is considered in this case, so this limits the contrast ratio in some part of the image. This results in significant contrast losses in small regions of the image, especially |

|   |   | in the background. |
|---|---|---|
| 3. | Local contrast enhancement technique is computationally complex. Sometimes, it involves high cost due to consideration of over-lapped sub-blocks. | The main advantage of global contrast enhancement is that it is computationally simple and is suitable for overall enhancement of the image. |
| 4. | Examples of two local contrast enhancement techniques that are frequently adopted are AHE and CLAHE. | Examples of two widely adopted Global contrast enhancement techniques are Histogram Equalization and Histogram Specification. |

In the following subsections, at first global contrast enhancement are discussed and then local contrast enhancement techniques with proper experiments showing their effect on satellite color images.

# II (1):GLOBAL CONTRAST ENHANCEMENT
## (A) HISTOGRAM EQUALIZATION

The histogram is a graph which shows the frequency of occurring of data in the whole data set. It plots the number of pixels for each tonal value (Bora,2016a)( Sutton). Consider an image with G total possible intensity levels. Then, the histogram of the image in [0, G-1] is defined as a discrete function:

$$p(r_k) = \frac{n_k}{n}$$

Where,

$r_k$ is the $k^{th}$ intensity level in the interval.

$n_k$ is the number of pixels in the image whose intensity level is $r_k$.

n is the total number of pixels in the image.

Histogram equalization is an image enhancement technique which enhances the contrast of an image by spreading the intensity values over the entire available dynamic range (Gonzalez et al,2014)(Bora et al, 2016a). This is achieved through a transformation function T(r), which can be defined by the Cumulative Distribution Function (CDF) of a given Probability Density Function (PDF) of gray levels in an image.

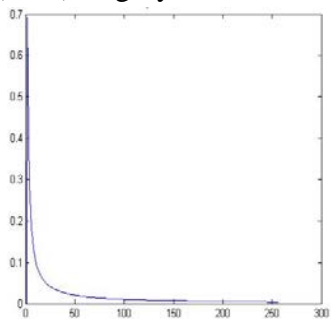

(a)

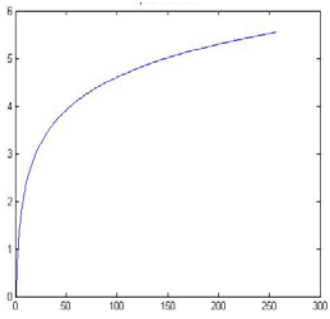
(b)

Fig1: (a) PDF; and (b) CDF

Here, we have two cases:

**(A) CONTINUOUS CASE:** This is for intensity levels that are continuous quantities normalized to the range [0, 1].

Let, $P_r(r)$ is the PDF of the intensity levels. Then, the required transformation on the input levels to obtain the output level S is:

$$S = T(r) = \int_0^r P_r(w)dw$$

where w is a dummy variable of integration. Then, it can be shown that (Gonzalez et al, 2014), the PDF of the output levels is uniform, i.e.,

$$P_s = \begin{cases} 1, & for\, 0 \leq s \leq 1 \\ 0, & otherwise \end{cases}$$

The above transformation generates an image whose intensity levels are equally likely and also, it covers the entire range [0, 1]. This intensity level equalization process results in an image with increased dynamic range with a tendency to have higher contrast.

**(B) DISCRETE CASE:** In the case of discrete quantities, we deal with summations and hence, the equalization transformation becomes:

$$S_k = T(r_k) = \sum_{j=1}^{k} P_r(r_j)$$

$$= \sum_{j=1}^{k} n_j/n \text{, for } k = 1, 2, 3, \ldots, L$$

where $S_k$ is the intensity value of the output image corresponding to value $r_k$ in the input image.

For our study, we have first converted the RGB image into HSV one, this is because HSV color space is more suitable for color image segmentation. This is due to the property that HSV color space represents the color in the same way that human eyes can perceive. Also, as we need the luminance channel, so we have extracted the V-channel from the HSV converted image and perform the histogram equalization operation on it. Finally, replacing the original V-channel with the histogram equalized one; we get the enhanced version of the image in HSV color space. So, with one more step (i.e., converting the HSV image back to RGB), we get the required histogram equalized version of the original satellite color image. Experimental results are shown below:

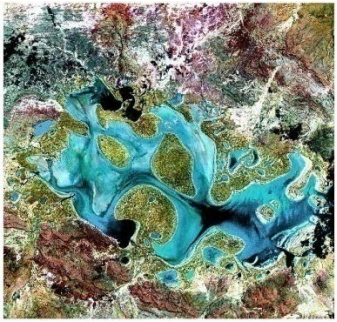
(a)

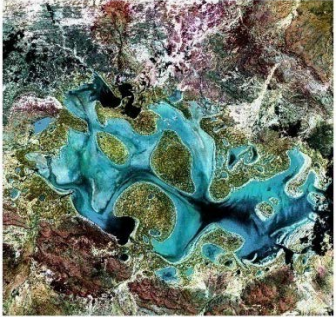
(b)

Fig 2: (a) Original Image and (b) Image obtained after histogram equalization.

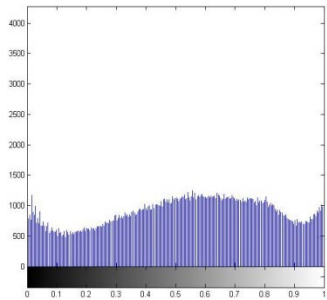
(c)

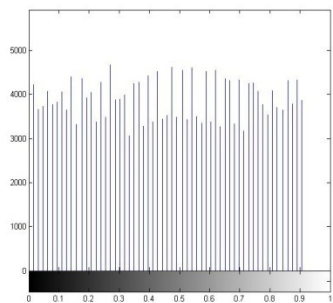
(d)

Fig 2(c): The Histogram of the V-channel of the Original Image, and (d) The Histogram of the V-channel of the Histogram Equalized Image

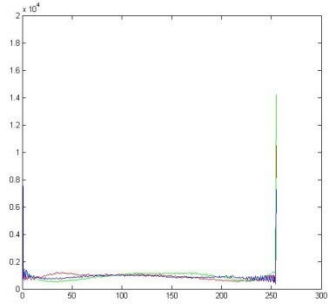
(e)

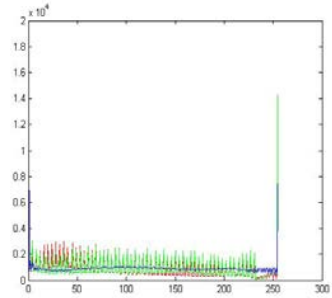
(f)

Fig 2: (e) Histogram of the original RGB image, and (f) Histogram of the image after histogram equalization.

## (B) HISTOGRAM SPECIFICATION

This is a technique by which the shape of the histogram of an image is changed into a specific one that we wish (Gonzalez et al, 2014) (Rajamani et al, 2013). Histogram specification is also known as Histogram Matching. Unlike the equally spaced ideal spaced histogram equalization, here the histogram is specified explicitly. If we consider continuous gray levels are normalized to the intervals [0,1] and say, r and z be the intensity levels of input and output image. Say, $p_r(r)$ is the input image's probability density function and $p_z(z)$ is the output ones (Gonzalez et al, 2014) Now, first apply the transformation:

$$s = T(r) = \int_0^r p_r(w)\, dw \qquad (1)$$

Above equation (1) will give an image with a uniform probability density. Then, if the desired output image were available, the following transformation would generate an image with uniform density:

$$H(z) = \int_0^z p_z(w)\, dw = s \qquad (2)$$

Where z denotes the intensity levels for an image that we are looking after. So, from equation (1) and equation (2), we have:

$$z = H^{-1}(s) = H^{-1}[T(r)]$$

So, a histogram specification technique consists of the following three main steps(Shaukat):

*Step1. Equalize the levels of the original image.*
*Step2. Specify the desired pdf and obtain the required transformation function (CDF).*
*Step3. Apply the inverse transformation function to the levels obtained in step 1.*

But, there exist a few drawbacks of histogram specification:
(1) Here, we do not have any definite rule for specifying an optimal histogram.
(2) Here, each given enhancement task needs to be analyzed on a case by case basis.
(3) Overall, this is a kind of trial-and-error process.

For experiment we have considered the same image used in the histogram equalization:

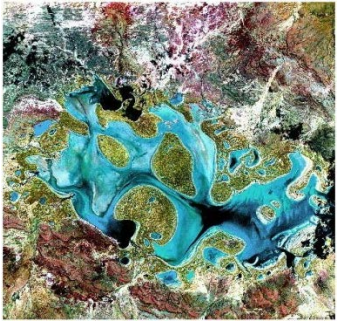

(a)

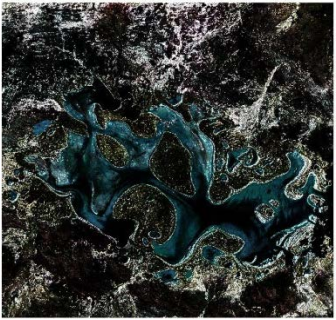

(b)

Fig 3: (a) Original Image and, (b) Image obtained after histogram specification.

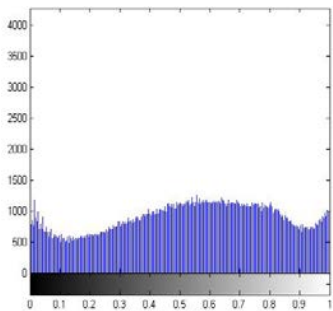

(c)

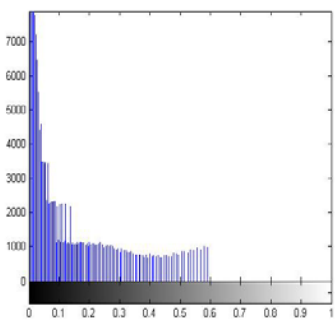

(d)

Fig 3(c): The Histogram of the V-channel of the Original Image, and (d) The Histogram of the V-channel of the Histogram Specified Image

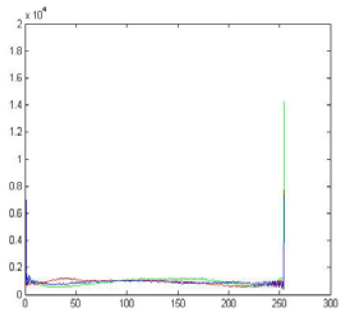
(e)

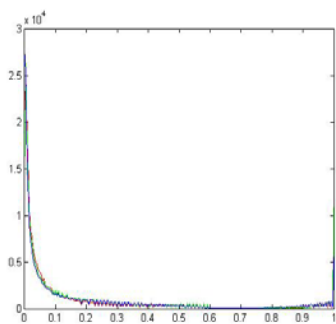
(f)

Fig 3: (e) Histogram of the original RGB image, and (f) Histogram of the image after histogram specification.

## II (2): LOCAL CONTRAST MANAGEMENT

### (A) AHE

AHE stands for Adaptive Histogram Equalization. But it is different from ordinary histogram equalization in the sense that it is not global and it computes many histograms corresponding to different sections of an image (Gupta et al,2014). So, it is possible to enhance the local contrast of an image through AHE. With AHE, the information of all intensity ranges of an image can be viewed simultaneously and thereby solving the problem of many ordinary devices which are unable to depict the full dynamic intensity range. Here, first, a contextual region is defined for every pixel in the image. The contextual region is the region centered about that particular pixel. Then, the intensity values for this region are used to find the histogram equalization mapping function. The mapping function thereby obtained is applied to the pixel being processed in the region and hence, the resultant image produced after each pixel in the image is mapping differently. This results in the local distribution of intensities and final enhancing are based on local area rather than the entire global area of the image. This is the main advantage of AHE. But, sometimes, AHE tends to over enhance the noise content that may exist in some homogeneous local block of the image by mapping a short range of pixels to a wide one.

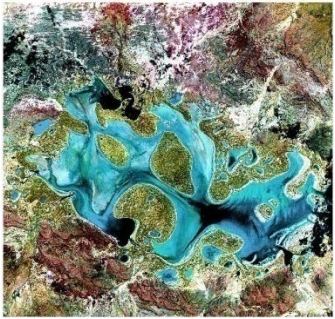
(a)

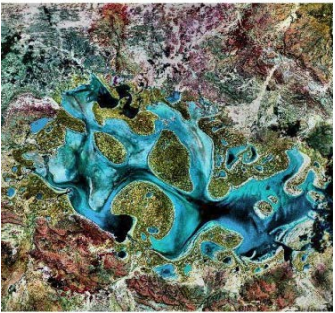
(b)

Fig 4: (a) Original Image and (b) Image obtained after AHE.

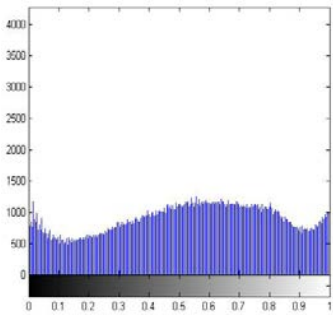
(c)

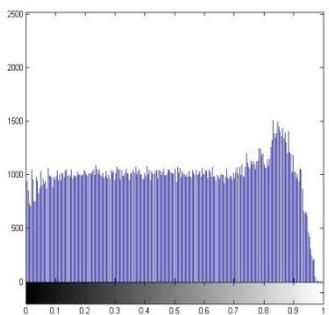
(d)

Fig 4: (c) The Histogram of the V-channel of the Original Image, and (d) The Histogram of the V-channel of the AHE enhanced Image

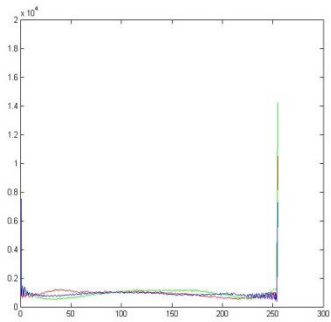

(e)

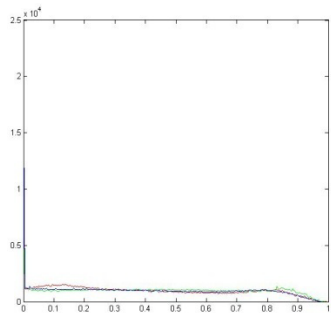

(f)

Fig 4: (e) Histogram of the original RGB image, and (f) Histogram of the image after AHE enhancement.

## (B) CLAHE:

CLAHE stands for Contrast Limited Adaptive Histogram Equalization. This is a local contrast enhancement technique, an improved version of AHE (Gupta et al, 2014) (Pizer et al,1987) (Bora,2016b). Actually, AHE suffers from amplification of noises that may exist in some homogenous regions, so, here arises the need of putting a limit to contrast enhancement. Through CLAHE, the input image is partitioned into some non-overlapping regions of equal size (approximate) and histogram equalization is applied on each of them (Gupta et al, 2014). Then every histogram is clipped by a clip limit which is based on the desired contrast expansion and the size of the neighboring region. After that, bilinear interpolation is applied to eliminate the region boundaries making them look smoother like as if no boundaries are there. CLAHE is based on the concept that if we put a limit or clip the height of the histogram of the concerned bin to a certain level then the slope of the CDF function can be limited automatically. This results in a limit on the local contrast enhancement to the desired extent. So, an efficient mechanism is needed to determine the points at which the clipping should be done and redistribute the clipped pixels. For this we have introduced a Binary Search Based CLAHE (BSB-CLAHE) (Bora et al, 2016b) whose algorithm is described below:

```
Step1. Say T is the top and B is the bottom of the concerned ClipLevel.
Step2. Repeat until T - B < ϵ, where ϵ > 0 is negligibly small
        [a] Find M where M is the middle between T and B
        [b] Find S where S is the sum of excess above M in each bin of the histogram
        [c] if S + M > ClipLevel
              then assign T = M
        [d] if S + M < ClipLevel
              then assign B = M
        [e] if S + M == ClipLevel
              then M is the required value at which clipping should be done. So, break binary search loop.
Step3. Clip the histogram at M and redistribute the excess into each bin equally.
```

After applying the BSB-CLAHE, we got the following results:

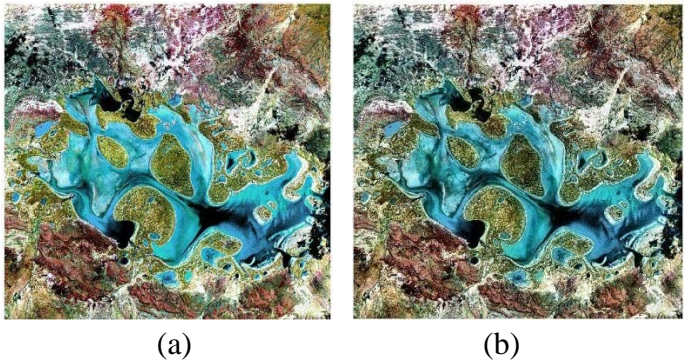

(a)            (b)

Fig 5: (a) Original Image and (b) Image obtained after histogram specification.

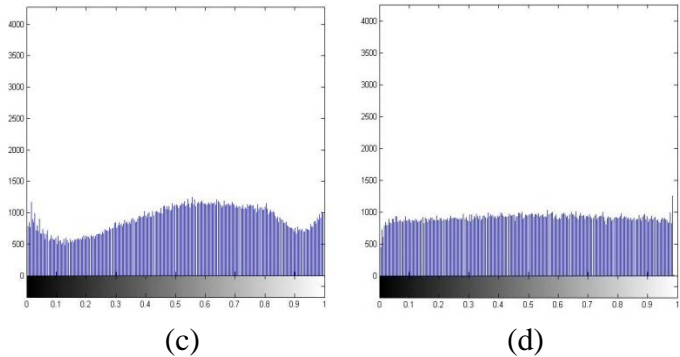

(c)            (d)

Fig 5: (c) The Histogram of the V-channel of the Original Image, and (d)The Histogram of the V-channel of the CLAHE enhanced Image

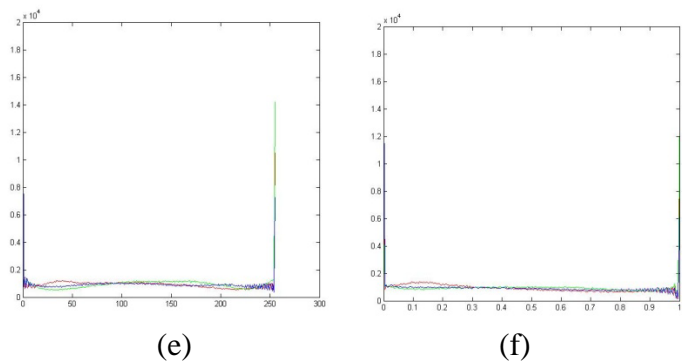

(e)            (f)

Fig 5: (e) Histogram of the original RGB image; and (f) Histogram of the image after CLAHE enhancement.

## III. COLOR IMAGE SEGMENTATION:

Image segmentation can be defined as dividing an image into different groups or partitions based on some homogeneity criteria like color, intensity, or texture etc.( Gonzalez et al, 2014) )(Bora et al, 2016b)(Gupta et al, 2016)(Ho,2011). This is a very important step in any image analysis process as it helps to identify the ROIs (Region of Interest) present in an image and hence catalyzes the further analysis process. Image segmentation is of two types: gray image segmentation and color image segmentation. Color image segmentation currently becomes a very emerging research area in image processing area. This is because of the reason that it has a better capability of isolating objects more distinctly than gray image segmentation. A gray image has only one channel associated with it, i.e. the intensity channel. While a color image has at least three channels associated with every image. As for example, in an RGB image, we have three channels R(Red), G(Green) and B(Blue) each carrying different information. So, a color image can bring much more information than a gray image. Also, it's also a fact that our human eyes are more adjustable to brightness, so, can identify thousands of color at any point of a complex image, while only a dozens of gray scale are possible to be identified at the same time(Bora et al, 2014a). This implies that color image segmentation has the ability to bring more information than gray image segmentation. For a better color image segmentation task, the first important issue is to decide which color space to be chosen for distribution of color attributes values of the concerned image. In many cases, it is found that HSV and L*A*B* are the two frequently adopted color spaces (Bora et al, 2014a)(Bora et al, 2014b)(Bora et al, 2016b). In the following figure 6(b) the gray segmentation of the image 6(a) is shown and in figure 6(c), the color image segmentation is shown. In both cases, we have adopted K-Means algorithm for the clustering task with Euclidean distance measure. Clustering is an unsupervised study commonly used for color image segmentation task. K-Means algorithm is a famous hard clustering algorithm popular for its low complexity (Bora et al, 2014c).

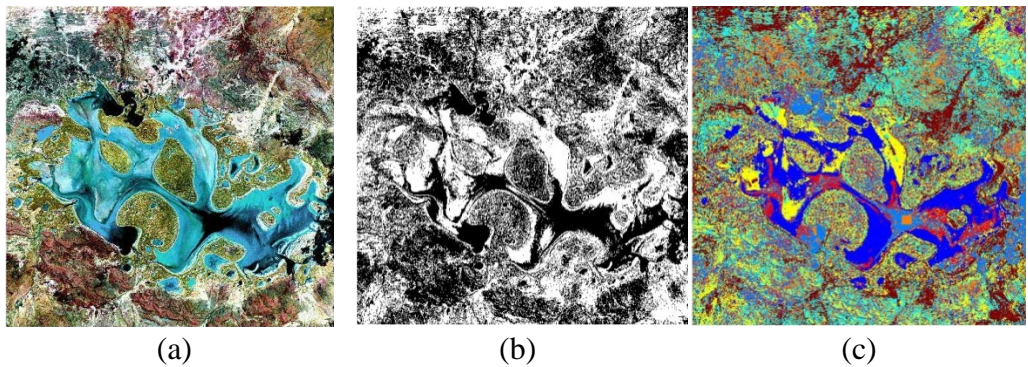

(a) (b) (c)

Fig 6: (a) Original Image; (b) Gray Image Segmentation; and (c) Color Image Segmentation.

From, the above figures it is clear we can recognize the ROI (Region of Interest) s more clearly in the image (c) than in (b), this is because of the reason that homogeneity in terms of color is more distinct and meaningful than the same in gray. Hence, color image segmentation can bring more valuable information than gray image segmentation.

## IV. COLOR SPACE:

A color space is an abstract mathematical model for which represents colors in terms of intensity values (Singha et al, 2011). It specifies how color information can be represented in combination with physical device profiling, thereby allowing us to understand the color capabilities of a particular device or digital file (2017). Color space can also be thought of just like a digital palette because it relates numbers to actual colors in three-dimensional coordinate system which contains all realizable color combinations ("Color Management: Understanding Color Spaces", 2017) Color space is the most important factor that needs to be considered first while going for any color image analysis process. There exist different types color spaces with respect to different types of applications and devices. Maximum times RGB is taken as a default color space in common sense. RGB defines a color as the percentage of red, green and blue hues mixed together (Poynton, 1995) ("Color Management: Understanding Color Spaces",2017).A color space can be either device dependent or device independent. Device dependent color space is somehow restricted to the parameters and device used for display. It may give a different display of colors in different devices. While in the case of device independent color spaces, we will get the same color regardless of whatever the device used. So, it is better to switch for device independent color spaces for color image analysis tasks like color image enhancement, color image segmentation. RGB is a device dependent color space. So, it is recommended that the enhancement process should not be done in this color space. HSV and LAB are two mostly preferred device independent color spaces. These two color spaces have a very strong and useful characteristic of separating 'luma' from 'chroma'. Here, luma refers to image intensity and chroma refers to color information. This characteristic is very important in image enhancement process as any enhancement needs to be done is on the luma channel only so that it will not has any adverse effect on the chroma channel otherwise it will result in the formation of strange irrelevant colors. A brief discussion on these two color spaces is given below.

## *HSV Color Space:*

In HSV color space, we have three channels: Hue (H), Saturation(S) and Value (V).Here, Hue is an angle in the range [0,2π] and is directly related to color. With respect to different hue angles, different colors will be presented. While saturation describes how pure the hue is with respect to a white reference and is measured as a radial distance from the central axis with values between 0 at the center to 1 at the outer surface. The V channel or value represents a percentage value goes from 0 to 100, expressing the amount of light illuminating a color. A diagrammatic view of the HSV color space is:

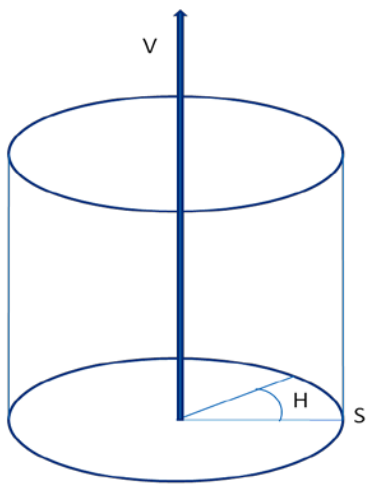

Fig. 7: HSV Color Space Showing the Relation between H, S, and V

The distribution of different colors with respect to Hue values (ranges 0 to 2π) is shown in the following color wheel diagram. Here, the color first started with red, as the Hue value changes,

then it changes in the following pattern:
red →yellow→green→cyan→blue→magenta→red.

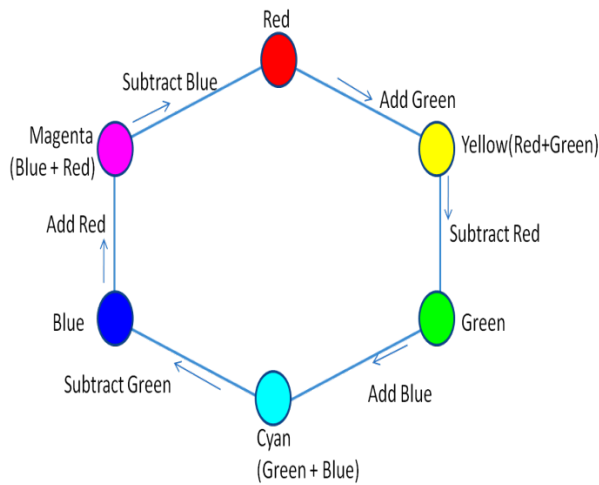

Fig. 8: Color Wheel Showing Distribution of Different Colors in HSV Color Space With Respect to Different Values of V.

## *LAB Color Space:*

LAB color space (L*A*B*) is a device independent color space defined by CAE and specified by the International Commission on Illumination(Bora et al,2014a)( Hunter et al, 1948)( Hunter et al, 1958). Here, L channel is for luminance or light, and the other layers a and b are chromaticity layers. The a* layer shows where the color falls along the red-green axis, and b* layer will indicate where the color will fall along the blue-yellow axis. A clear illustration of the coordinate system of lab color space can be presented with the following diagram(2017):

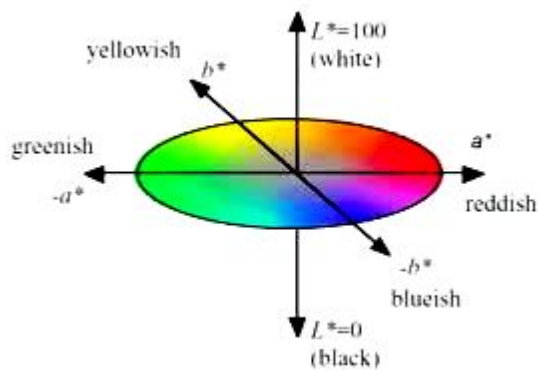

Fig. 9: LAB Color Space

So, from the figure (9), it is clear that the a* negative values indicate green while positive values indicate magenta; and b* negative values indicate blue and positive values indicate yellow.

For, our study we will perform all the enhancement techniques on both of these two color spaces.

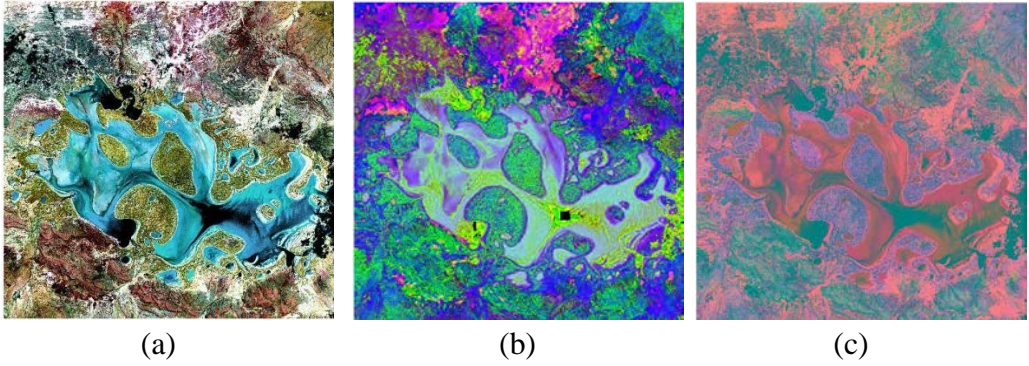

(a) (b) (c)

Fig. 10: (a) Original Image; (b) Image in HSV Color Space; and (c) Image in LAB Color Space.

## V. EXPERIMENTS AND DISCUSSION:

For the experimental analysis, we have chosen Matlab to implement the enhancement techniques. The system where experiments are done has i5 processor with 64 bit Windows 10 operating system. Color Images required for the experiments are collected from different sources like Berkeley Image Segmentation Dataset, Sun Database, Earth Science World Image Bank. For this comparative study, we will first enhance the images with global contrast enhancement techniques and then with local contrast enhancement techniques. Then the resultant image will be undergone color image segmentation with an efficient clustering-based technique (Bora et al,2016b). Then we will compare the results and thereby come to the conclusion that which enhancement technique will be more suitable for color image segmentation.

# 1. ENHANCEMENT IN HSV COLOR SPACE:

| | Original Image | Histogram Equalization | Histogram Matching | AHE | BSB-CLAHE |
|---|---|---|---|---|---|
| 1. Lena Image | 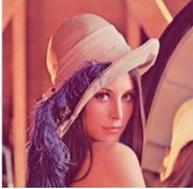 | 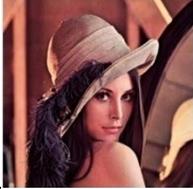 | 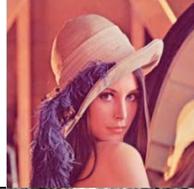 | 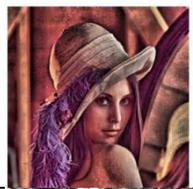 | 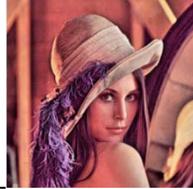 |
| 2. Beer (Over Brightness) | 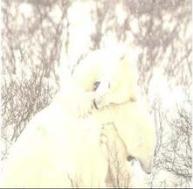 | 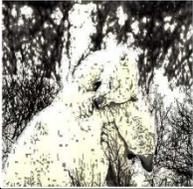 | 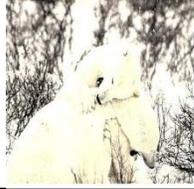 | 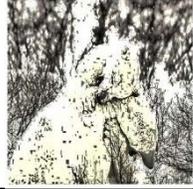 | 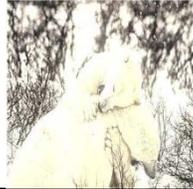 |
| 3. Underwater Image | 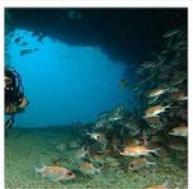 | 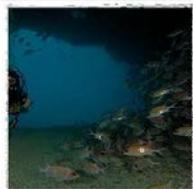 | 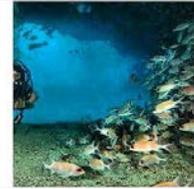 | 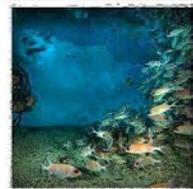 | 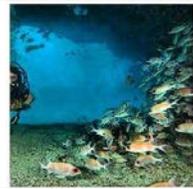 |
| 4. Satellite Blurred Image | 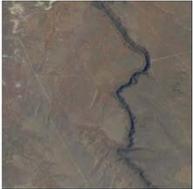 | 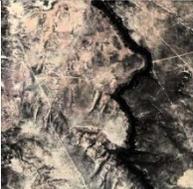 | 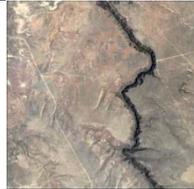 | 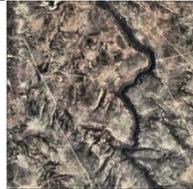 | 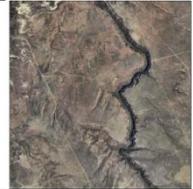 |
| 5. Noisy Face Image | 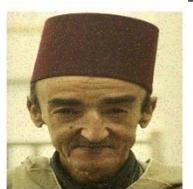 | 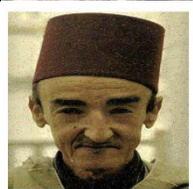 | 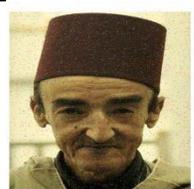 | 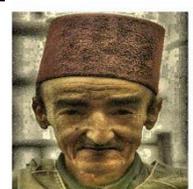 | 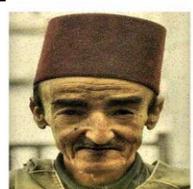 |

Fig. 11: (a) Original Image; (b) Enhancement by Histogram Equalization; (c) Histogram Matching ; (d)AHE; (e) BSB-CLAHE.

## 2. ENHANCEMENT IN LAB COLOR SPACE

| | Original Image | Histogram Equalization | Histogram Matching | AHE | BSB-CLAHE |
|---|---|---|---|---|---|
| 1. Lena Image | 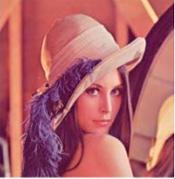 | 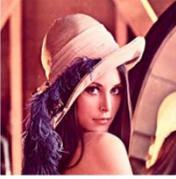 | 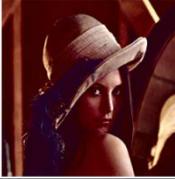 | 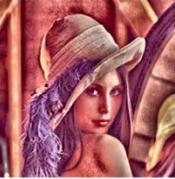 | 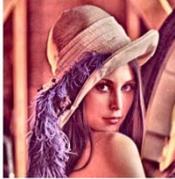 |
| 2. Beer (Over Brightness) | 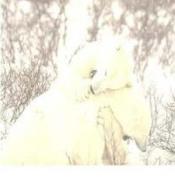 | 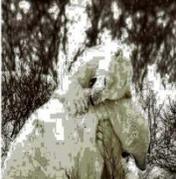 | 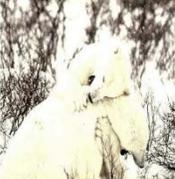 | 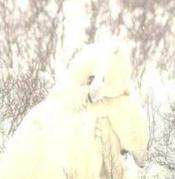 | 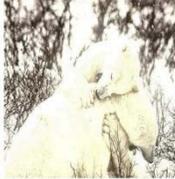 |
| 3. Underwater Image | 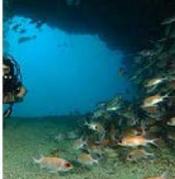 | 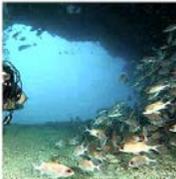 | 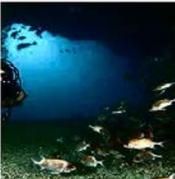 | 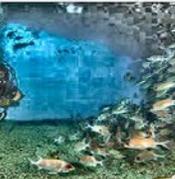 | 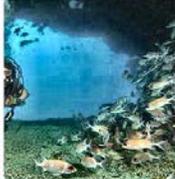 |
| 4. Satellite Blurred Image | 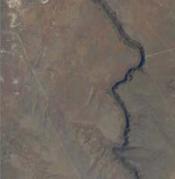 | 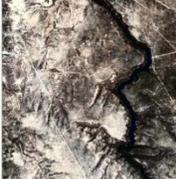 | 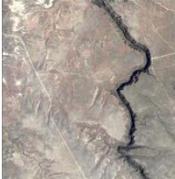 | 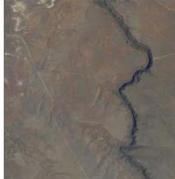 | 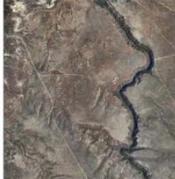 |
| 5. Noisy Face Image | 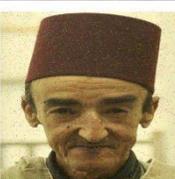 | 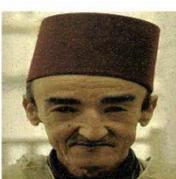 | 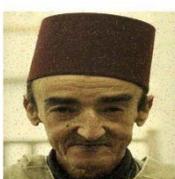 | 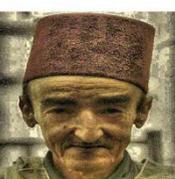 | 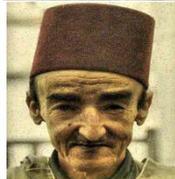 |

Fig. 12: (a) Original Image; (b) Enhancement by Histogram Equalization; (c) Histogram Matching ; (d)AHE; (e) BSB-CLAHE.

For the above experiments, we have selected five different images: Lena(regarded as standard test image), beer image(with over brightness), underwater image, satellite image(blurred one) and a noisy face image(with salt & pepper noise of density 0.2 added). These images are undergone enhancement with each technique discussed in this paper. The enhancements are done on two color spaces separately: HSV and LAB. Here, we are using two metrics for quality for quality measurement: first visual perspective and then entropy. As per enhancement is concerned visual perspective should be given first preference. Then the minute differences are brought into light through the entropy value calculation. A high value of entropy implies a large amount of information stored in the image (Iyad et al,2007)( Pal et al, 1991). So, a higher value of entropy means a better enhancement is carried out. Entropy, E can be calculated by using the following equation:

$$E = -sum(p.\log_2(p))$$  (1)

where p is the histogram counts involved in the histogram of the concerned image. As we have a color image, so, for entropy calculation, it is taken as a multidimensional gray image.

Table 1: Entropy Values Comparison

|  | Color Space | Histogram Equalization | Histogram Matching | AHE | BSB-CLAHE |
|---|---|---|---|---|---|
| Lena | HSV | 6.1331 | 6.1269 | 6.0696 | 6.2568 |
|  | LAB | 5.9683 | 5.2271 | 6.1102 | 6.0851 |
| Beer | HSV | 4.5560 | 4.4827 | 4.8412 | 4.4856 |
|  | LAB | 5.4472 | 4.9086 | 4.0502 | 4.9316 |
| Underwater Image | HSV | 2.7851 | 3.0319 | 3.6792 | 3.0319 |
|  | LAB | 4.6273 | 4.0595 | 4.7340 | 4.7427 |
| Satellite Image | HSV | 6.1617 | 5.5729 | 6.0606 | 5.5090 |
|  | LAB | 6.1470 | 5.5362 | 4.8225 | 5.5153 |
| Noisy Face Image | HSV | 5.5523 | 5.4807 | 5.4597 | 5.5494 |
|  | LAB | 5.5069 | 4.7367 | 5.4546 | 5.5382 |

Now, from the above experimental results, it is seen that enhancement on HSV color space is better for every enhancement techniques than the same in LAB color space. Also, as per visual perspective is concerned, BSB-CLAHE is showing best performance on both the color spaces and for all the images. After that, on average histogram matching is showing good performance. While, we have seen that although AHE is showing better performance than histogram equalization (global), but sometimes it fails to produce desired output. Histogram equalization is found often suffered from over enhancement issue-producing unwanted dark regions or artifacts. So, it is not recommended to use histogram equalization as a pre-processing step in any image analysis process. But, if we talk about computational complexity, the histogram equalization is less complex, while BSB-CLAHE is highly complex. While for histogram specification, the complexity will totally dependent on the specified histogram. Histogram specification is better only if we have depth background knowledge of the concerned image data.

Now, in the next stage of our experimental, we will analyze the performance of the discussed enhancement techniques with respect to color image segmentation. For segmentation purpose, an approach AERASCIS(Bora et al,2016b) is adopted as its performance is better than the traditional algorithms. The segmentation is done on both the color spaces for several test images collected from the above-mentioned sources. Following is the output for peppers image which is frequently selected for analyzing the performance of segmentation algorithms as in this image from visual perspective there is distinct ROI s present in different colors. Segmentation is done on both HSV and LAB color spaces.

Following flowchart shows the steps involved in the whole process:

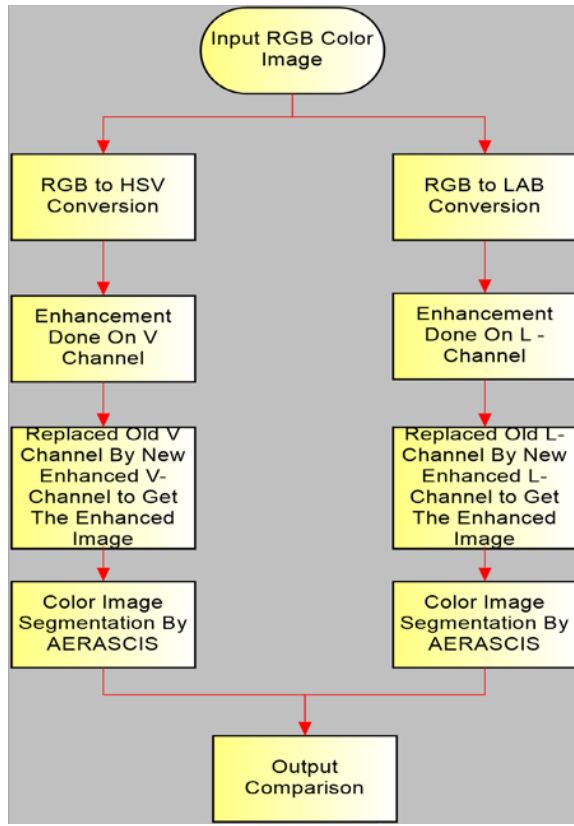

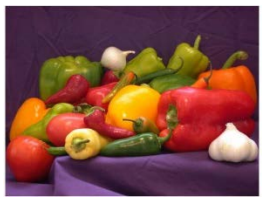

Original Peppers Image

| Color Space | Histogram Equalization Based Segmentation | Histogram Specification Based Segmentation | AHE Based Segmentation | CLAHE Based Segmentation |
|---|---|---|---|---|
| HSV | 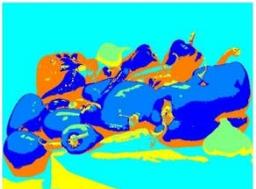 | 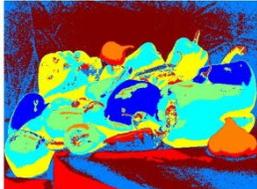 | 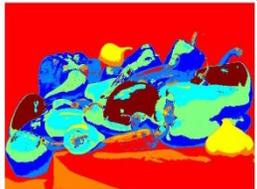 | 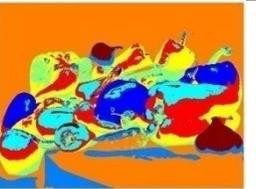 |
| LAB | 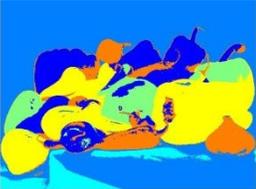 | 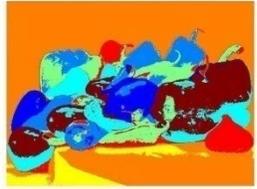 | 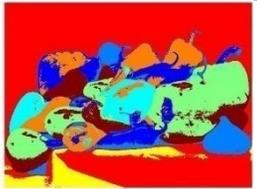 | 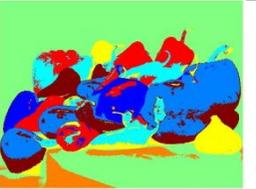 |

The 'peppers' image is now added a Gaussian noise of mean 0 and variance 0.01 to find the robustness of the segmentation with respect to noise.

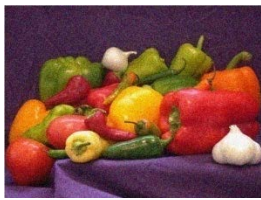 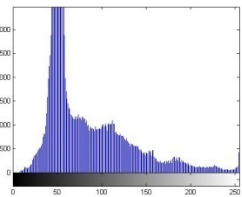

(a)Noisy Image          (b)Original Histogram

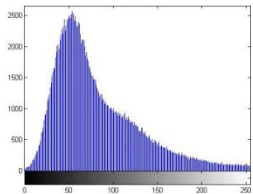

(c)Histogram for the Noisy Version

Then the noisy image is undergone segmentation with preprocessing of the discussed enhancement techniques separately and results are mentioned below:

| Color Space | Histogram Equalization Based Segmentation | Histogram Specification Based Segmentation | AHE Based Segmentation | CLAHE Based Segmentation |
|---|---|---|---|---|
| HSV | 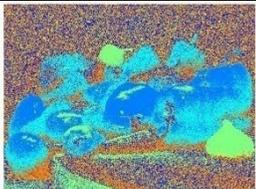 | 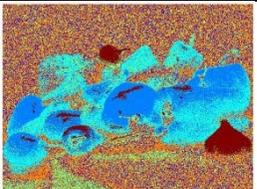 | 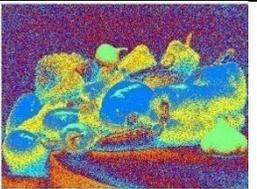 | 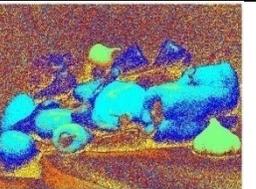 |
| LAB | 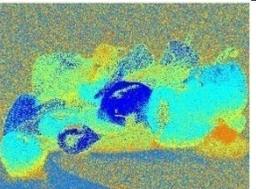 | 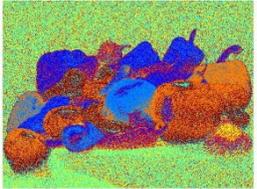 | 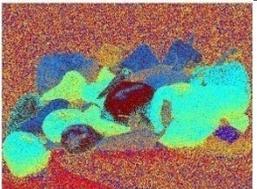 | 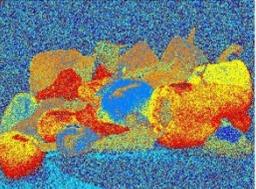 |

Table 2: MSSIM values Comparison

| Segmented Under Condition | Color Space | Histogram Equalization | Histogram Specification | AHE | BSB-CLAHE |
|---|---|---|---|---|---|
| Without Gaussian Noise | HSV | 0.4948 | 0.4952 | 0.4950 | 0.4967 |
|  | LAB | 0.4132 | 0.4613 | 0.4510 | 0.4719 |
| With Gaussian Noise | HSV | 0.4046 | 0.4438 | 0.4419 | 0.4504 |
|  | LAB | 0.3482 | 0.3463 | 0.3453 | 0.4100 |

Now, from the above results we can come to the following conclusions by considering the visual perspective point of view:

1. Without Gaussian Noise: it is clear that segmentation after preprocessing with BSB-CLAHE is giving the best performance in both the color spaces. After that, histogram specification is giving better performance than AHE and Histogram Equalization. It is seen that the segmentation algorithm is producing better segmentation with respect to AHE than histogram equalization.

2. With Gaussian Noise: When the segmentation algorithm is applied to segment the Gaussian noise added peppers image after preprocessing with the discussed enhanced techniques on both the color spaces separately, it is found that LAB color space fails to deal with the noisy color image even after the enhancement is done. So, it draws the fact that LAB is not good for noisy image enhancement and segmentation. Only HSV color space survives to the noise and succeeds to produce considerable segmentation when the preprocessing is done with the discussed enhancement techniques. In this case, also BSB-CLAHE is able to give the best performance. While Histogram Specification and AHE are showing the average performance, histogram equalization is the one with the worst performance.

Now, after the analysis with respect to visual perspective, we go for quantitative evaluation of the results for mathematical proof of quality. For that task, we have adopted SSIM index proposed by Zhou Wang et al. (Wang et al, 2004). Usually, researchers use to take MSE for quality analysis but there are several cases, it does not give an accurate analysis of the results (Wang et al, 2009). SSIM has the ability to automatically predict perceived image quality. A detail on this structural Structural-Similarity-Based Image Quality Assessment can be found in (Wang et al, 2004). We have used mean SSIM (MSSIM) index to estimate the overall image quality:

$$mssim(IMG1, IMG2) = \frac{1}{N} \sum_{i=1}^{N} ssim(img1_i, img2_i)$$

where IMG1 is the original image and IMG2 is its segmented version.; $img1_i$ and $img2_i$ are the image contents in the ith local window and N is the total number of the local windows of the image. The MSSIM values are shown in the above table 2.

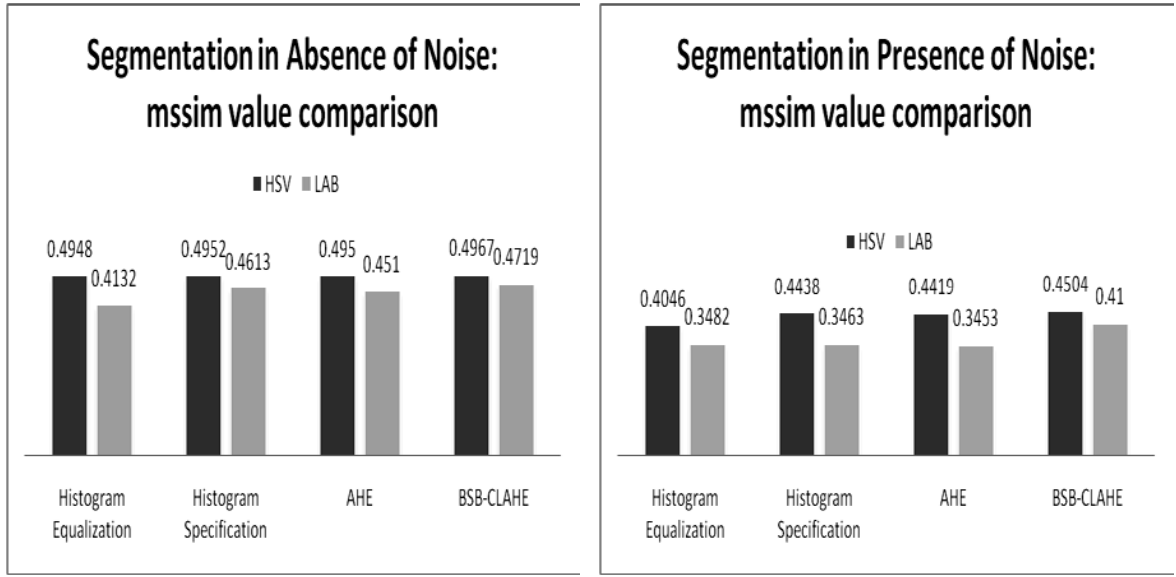

So, from the above table 2, it is seen that BSB-CLAHE based color image segmentation results in higher mssim values than the same of other enhancement techniques based segmentation. A higher value of mssim implies a better quality of segmentation. Also, the segmentation results are better in HSV color space than the LAB color space. Now, if we arrange the enhancement technique on the basis of their performance then it follows the following order (here '>' means better than ):

***BSB-CLAHE > Histogram Specification > AHE> Histogram Equalization.***

So, CLAHE based local image enhancement technique is the best option for color image enhancement as it is robust to noise as we have seen in the experiments that it succeeds to produce better results even in the presence of frequently found noises: Salt and Pepper and Gaussian. As a result, it overall increases the performance of color image segmentation.

## VI. CONCLUSION

Image pre-processing is utmost required for a better image analysis. In the case of color image processing, color image enhancement is the mostly concerned topic as distortion in the color image will impact the later analysis process like segmentation very negatively and hence this distortion should be removed through enhancement techniques as much as possible. Contrast enhancement is one of such preprocessing techniques that are frequently adopted. Various contrast enhancement techniques exist which are classified mainly into two types: local and global. In this paper, a comprehensive and comparative study is done on such contrast enhancement techniques and their impact on color image segmentation. Four techniques histogram equalization, histogram specification, AHE and BSB-CLAHE are discussed in details. Here color space is also given major concern. All the experiments are done HSV and LAB color space separately. It is found that HSV color space is showing better performance than LAB color space even in presence of noises for color image enhancement as well as color image segmentation. After analyzing all the experimental results, we come to the conclusion that BSB-CLAHE as a preprocessing technique will bring better color image segmentation if the task is performed in HSV color space. So, any CLAHE based local contrast enhancement technique (where clip limit is determined properly) on HSV color space should be preferred than other global enhancement techniques like histogram equalization if we want to achieve an optimal color image segmentation result.

# REFERENCES

(2017). Retrieved 7 January 2017, from http://en.wikipedia.org/wiki/Color_space.

Bora, D., & Gupta, A. (2016b). AERASCIS: An efficient and robust approach for satellite color image segmentation. 2016 International Conference On Electrical Power And Energy Systems (ICEPES). http://dx.doi.org/10.1109/icepes.2016.7915989.

Bora, D., & Gupta, D. (2014c). A Comparative study Between Fuzzy Clustering Algorithm and Hard Clustering Algorithm. International Journal Of Computer Trends And Technology, 10(2), 108-113. http://dx.doi.org/10.14445/22312803/ijctt-v10p119.

Bora, D.J., & Kumar Gupta, A. (2014b). A New Approach towards Clustering based Color Image Segmentation. International Journal Of Computer Applications, 107(12), 23-30. http://dx.doi.org/10.5120/18803-0329.

Bora, D.J., Gupta, A.K. (2014a). A Novel Approach Towards Clustering Based Image Segmentation. *International Journal of Emerging Science and Engineering (IJESE)*, ISSN: 2319–6378, 2(11), 6-10.

Bora, D.J., Gupta, A.K., (2016a). A New Efficient Color Image Segmentation Approach Based on Combination of Histogram Equalization with Watershed Algorithm. International Journal of Computer Sciences and Engineering, 4(6), 156-167.

Color Management: Understanding Color Spaces. (2017). Cambridgeincolour.com. Retrieved 30 January 2017, from http://www.cambridgeincolour.com/tutorials/color-spaces.htm.

Daytime Color Appearance of Retro reflective Traffic Control Sign Materials. (2017). Federal Highway Administration Research and Technology Coordinating, Developing, and Delivering Highway Transportation Innovations. Retrieved 13 January 2017, from https://www.fhwa.dot.gov/publications/research/safety/13018/13018.pdf.

Earth Science World Image Bank. (n.d.). Retrieved from http://www.earthscienceworld.org/imagebank/

Gonzalez, R., & Woods, R. (2014). *Digital image processing* (1st ed.). New Delhi: Dorling Kindersley.

Gupta, A.K. , Bora, D.J.(2016). A Novel Color Image Segmentation Approach Based On K-Means Clustering with Proper Determination of Number of Clusters and Suitable Distance Metric. *International Journal of Computer Science & Engineering Technology (IJCSET)*,ISSN :2229-3345, 7(9), 395-409.

Gupta, S., & Kaur, Y. (2014). Review of Different Local and Global Contrast Enhancement Techniques for a Digital Image. International Journal Of Computer Applications, 100(18), 18-23. http://dx.doi.org/10.5120/17625-8384

Ho , Pei-Gee.(2011). Image Segmentation. Book, Published By InTech Janeza Trdine 9, 51000 Rijeka, Croatia.

Hunter, R. (1948). Accuracy, precision, and stability of new photo-electric colordifference meter. Josa 38 (12): 1094. (Proceedings of the thirty-third annual meeting of the optical society of America).

Hunter, R. (1958). Photoelectric Color Difference Meter*. Journal Of The Optical Society Of America, 48(12), 985. http://dx.doi.org/10.1364/josa.48.000985.

Iyad Jafar, & Hao Ying. (2007). Multilevel component-based histogram equalization for enhancing the quality of grayscale images. 2007 IEEE International Conference on Electro/Information Technology. doi:10.1109/eit.2007.4374490.

Koschan, A., & Abidi, M. (2008). Digital color image processing. Hoboken, NJ: Wiley-Interscience.

Malik, S., & Lone, T. (2014). Comparative study of digital image enhancement approaches. 2014 International Conference On Computer Communication And Informatics. http://dx.doi.org/10.1109/iccci.2014.6921749.

Pal, N., & Pal, S. (1991). Entropy: a new definition and its applications. IEEE Transactions on Systems, Man, and Cybernetics, 21(5), 1260-1270. doi:10.1109/21.120079.

Pizer, S., Amburn, E., Austin, J., Cromartie, R., Geselowitz, A., & Greer, T. et al. (1987). Adaptive histogram equalization and its variations. Computer Vision, Graphics, And Image Processing, 39(3), 355-368. http://dx.doi.org/10.1016/s0734-189x(87)80186-x.

Poynton, C. (1995). A Guided Tour of Colour Space. New Foundation For Video Technology: The SMPTE Advanced Television And Electronic Imaging Conference. http://dx.doi.org/10.5594/m00840.

Rajamani, V.; Babu , P.; Jaiganesh, S. (2013). A Review of various Global Contrast Enhancement Techniques for still Images using Histogram Modification Framework. International Journal of Engineering Trends and Technology (IJETT). V4(4):1045-1048.

Shaukat,A. Histogram Processing. Web link: https://sgar91.files.wordpress.com/2011/04/dip_lecture_8.pdf.

Singha ,M., Hemachandran,K.(2011). Performance analysis of Color Spaces in Image Retrieval. Assam University Journal of science & Technology, 7( II), 94-104.

Sutton, Ed. Histograms and the Zone System. Illustrated Photography.

The Berkeley Segmentation Dataset and Benchmark. (2017). www2.eecs.berkeley.edu. Retrieved 15 January 2017, from https://www2.eecs.berkeley.edu/Research/Projects/CS/vision/bsds/

Understanding Color Spaces and Color Space Conversion - MATLAB & Simulink - MathWorks United Kingdom. (2017). In.mathworks.com. Retrieved 9 January 2017, from http://in.mathworks.com/help/images/understanding-color-spaces-and-color-space-conversion.html?refresh=true.

Wang, Z. & Bovik, A. (2009). Mean squared error: Love it or leave it? A new look at Signal Fidelity Measures. IEEE Signal Processing Magazine, 26(1), 98-117. doi:10.1109/msp.2008.930649

Wang, Z., Bovik, A., Sheikh, H., & Simoncelli, E. (2004). Image Quality Assessment: From Error Visibility to Structural Similarity. IEEE Transactions on Image Processing, 13(4), 600-

612. doi:10.1109/tip.2003.819861

Xiao, J., Hays, J., Ehinger, K. A., Oliva, A., & Torralba, A. (2010). SUN database: Large-scale scene recognition from abbey to zoo. 2010 IEEE Computer Society Conference on Computer Vision and Pattern Recognition. doi:10.1109/cvpr.2010.5539970.

ResearchGate Link:

https://www.researchgate.net/publication/318901891_IMPORTANCE_OF_IMAGE_ENHANCEMENT_TECHNIQUES_IN_COLOR_IMAGE_SEGMENTATION_A_COMPREHENSIVE_AND_COMPARATIVE_STUDY